\documentclass{ecai}
\usepackage{times}
\usepackage{graphicx}
\usepackage{latexsym}
\usepackage{soul}
\usepackage{comment}
\usepackage{amsmath}
\usepackage{enumitem}
\usepackage{xcolor}
\usepackage[linesnumbered, ruled, lined, boxed]{algorithm2e}
\usepackage[
singlelinecheck=false 
]{caption}
\makeatletter
\newcommand{\removelatexerror}{\let\@latex@error\@gobble}
\makeatother

\begin{document}

\title{
Prognosis Prediction in Covid-19 Patients from Lab Tests and X-ray Data
through Randomized Decision Trees}
\author{
  Alfonso E. Gerevini$^*$, Roberto Maroldi$^{*\dagger}$, Matteo Olivato$^*$, Luca Putelli$^*$, Ivan Serina$^*$\\
  $^*$Università degli Studi di Brescia, $^\dagger$ASST Spedali Civili di Brescia\\
 \{alfonso.gerevini, roberto.maroldi, ivan.serina, m.olivato, l.putelli002\}@unibs.it
  }
 
\maketitle
\bibliographystyle{ecai}

\begin{abstract}
AI and Machine Learning can offer powerful tools to help in the fight against  Covid-19. In this paper we present a study and a concrete tool based on machine learning to predict the prognosis of hospitalised patients with Covid-19. In particular we address the task of predicting the risk of death of a patient at different times of the hospitalisation, on the base of some demographic information, chest X-ray scores and several laboratory findings. Our machine learning models use ensembles of decision trees trained and tested using data from more than 2000 patients. An experimental evaluation of the models shows good performance in solving the addressed task.\footnote{This paper is published in Proceedings of 5th International Workshop on Knowledge Discovery in Healthcare Data (KDH) at ECAI 2020.}

\end{abstract}
\vspace{-5mm}
\section{Introduction}
The fight against Covid-19 is a new important challenge for the world that AI and machine learning can help facing at various levels \cite{jiang, van2020artificial, yan2020prediction}. In March 2020, at the time of the coronavirus emergency in Italy, we started working in strict collaboration with one of the hospitals that had more Covid-19 patients in Italy, \textit{Spedali Civili di Brescia}, to help predicting the prognosis of hospitalised patients.  Our work was focused on the task of predicting the risk of death of a patient at different times of the hospitalisation.
As discussed in \cite{van2020artificial},
predicting if a patient is at risk of decease or adverse events can help the hospital, for instance, to organize the allocation of limited health resources in a more efficient way.

Our predictive models are built
on the base of demographic information (sex and age), the values of ten laboratory tests and the chest X-ray score(s), which is an innovative measure developed and used at \textit{Spedali Civili di Brescia}
to assess the severity of the pulmonary conditions \cite{maroldi_score}.
Other important information, such us the patient comorbidities or the time and duration of the symptoms related to Covid-19, were not used because not available to us.
%

Using raw data from more than 2000 patients, we built some data sets describing the ``clinical history'' of each patient during the hospitalisation.
In particular, each dataset contains a ``snapshot'' of the infection conditions of every considered patient at a certain day after the start of the hospitalisation. For each dataset, we built a different predictor, allowing to make progressive predictions over
time that take into account the evolution
of the disease severity in a patient, which helps the formulation of a personalized prediction of the prognosis. A change of the predicted risk over time for a patient could also hint a link between specific events or treatments and the increase or decrease of the risk for the patient.
As snapshot times for a patient, in our experiments we considered the 2nd, 4th, 6th, 8th and 10th hospitalization day, and the day before the end of the hospitalisation.

Our datasets were engineered to cope with a number of practical issues, including
missing values and feature values categorization,  and to add some helpful artificial features. We also  addressed the ``concept drift'' issue \cite{concept_drift,dataset_shift}, since we observed that the risk of death was clearly sensitive  to the time period when the patient was hospitalised; the risk was significantly higher during the earlier period of the emergency (March 2020), when in northern Italy the spread of the virus infection was very high and many people were hospitalised.
%
Moreover, given the very sensitive nature of our task, we introduced a threshold to discharge the model predictions that have a low estimated probability. Such a threshold is a parameter that is automatically 
calculated 
and optimised during the training phase.

We considered several machine learning algorithms. A first experimental comparison of their performance on our data sets showed that methods based on forests of trees have more promising performance, and so we decided to focus on this approach. 
The obtained prediction models have good performance over a randomly chosen test set of 200 patients 
for each considered period,
in terms of both F2 and ROC-AUC scores. In particular, overall the system makes very few errors in predicting patient survival, i.e., the specificity of the prediction is very high. 

In the following, after discussing related work, we describe our data sets, we present our prediction models and their experimental evaluation, and finally we give conclusions and mention future work.
\vspace{-5mm}
\section{Related work}
Artificial Intelligence and Machine Learning techniques can be used for tackling the Covid-19 pandemic in different aspects. However, given that the pandemic has started only few months ago, most works are still preliminary, and there isn't a clear description of the developed techniques and of their results (often only pre-printed and not properly peer-reviewed). 

A preliminary study is presented in \cite{jiang}. Given a set of only 53 patients with mild symptoms and their lab tests, comorbidities and treatment, the authors train several machine learning models (Logistic Regression, Decision Trees, Random Forests, Support Vector Machines, KNN) to predict if a patient will be subject to more sever symptoms, obtaining a prediction accuracy score of up to 0.8 using 10-fold cross validation. The generalizability and strength of these results are questionable, given the very small set of considered patients.
Another example is the pre-printed work by Li Yan {\it et al.} \cite{yan2020prediction} that uses lab tests for predicting the mortality risk; the proposed model is a very simple decision tree based on the three most important features. While the performance seems promising, the test set used for evaluation was very small (29 patients).

Various AI and machine learning techniques have been developed for prognosis and disease progression prediction \cite{GereviniAIME} in the context of diseases different from Covid-19 \cite{kes_2020, Putelli2019TheIO, putelli_self}.
In particular, in the last few years, several works about predicting mortality risk or adverse events and on the use of AI in critical care
\cite{pollard2017enabling} have been published. 
The survey in \cite{mortality_ml} presents a review of statistical and ML systems for predicting the mortality risk, the need of beds in intense care units \cite{yoon2016forecasticu} or the length of the patient hospitalization.
In particular, it is worth to mention the work by Harutyunyan et al. \cite{harutyunyan2017multitask} which uses LSTM Neural Networks for predicting both the mortality risk and the length of the hospitalisation.

An overview of the issues and challenges for applying ML in a critical-care context is available in \cite{johnson2016machine}. This work stresses the need to deal with  corrupted data, like missing values, imprecision, and errors that can increase the complexity of prediction tasks.

Lab test findings and their variation over time are the main focus of the work by Hyland et al. \cite{hyland}, which describes a system that processes these data to generate
an alarm predicting that a patient will have a circulatory failure 2 hours in advance.

\section{Available Data Sources}
During the Covid-19 outbreak, from February to April 2020 in hospital \textit{Spedali Civili di Brescia} more than two thousand patients were hospitalised. 
During their hospitalisation, the medical staff performed several exams to them in order to monitor their conditions, checking the response to some treatments, verifying the need to transfer a patient to the ICU, etc.
We had data from a total of 2015 hospitalised patients; for each of these patients, the specific data that were made available to us are: 
\begin{itemize}
    \item the age and sex;
    \item the values and dates of several lab tests (see Table \ref{tab:lab_values});
    \item the scores (each one from $0$ to $18$), assigned by the physicians, assessing the severity of the pulmonary conditions resulting from the X-ray exams \cite{maroldi_score};
    \item the values and dates of the throat-swab exams for Covid-19;
    \item the final outcome of the hospitalisation at the end of the stay, which is the classification value of our application (either in-hospital death, released survivor, or transferred to another hospital or rehabilitation center).
\end{itemize}

Table \ref{tab:lab_values} specifies the considered lab tests, their normal range of values, and their median values in our set of patients.
We had no further information about symptoms, their timing, comorbidities, generic health conditions or clinical treatment. Moreover, we have no CT images or text reports associated with the X-ray exams. 
The available information about whether a patient was or had been in ICU was not clear enough to be used. 
Finally, of course, also the names of the patient and of the involved medical staff names were not provided. 

\subsection{Data Quality Issues}
\label{sec:issues}
\begin{table}
    \begin{tabular}{l| c |c}
    Lab test & Normal Range & Median Value \\ \hline
    C-Reactive Protein (PCR) & $\leq 10$ & 34.3 \\
    Lactate dehydrogenase (LDH) & [80, 300] & 280 \\
    Ferritin (Male) & [30, 400] & 1030 \\
    Ferritin (Female) & [13, 150] & 497 \\
    Troponin-T & $\leq 14$ & 19 \\
    White blood cell (WBC) & [4, 11] & 7.1 \\
    D-dimer & $\leq 250$ & 553 \\
    Fibrinogen & [180, 430] & 442 \\
    Lymphocite (over 18 years old patients) & [20, 45] & 1.0 \\
    Neutrophils/Lymphocites & [0.8, 3.5] & 4.9 \\
    Chest XRay-Score (RX) & $<7$ & 8 \\
    
    \end{tabular}
    \caption{Lab tests performed during the hospitalisation. In the second column, we show the range which is considered clinically normal for a specific exam. In the third column, we show the median value extracted considering the lab test findings for our set of 2015 patients.}
    \label{tab:lab_values}
\end{table}
\label{sec:issues}
When applying machine learning to raw real-world data, there are some non-trivial practical issues to deal with, such as the quality of the available data and related aspects, that in biomedical applications are especially important given the very sensitive domain \cite{hasan2006analyzing}.

In our case, one of such issues is that the length of the hospitalisation period can sensibly differ from one patient to another (from few days to two months), due to different reasons including
the novelty and the characteristics of the disease, its high contagiousness or the absence of an effective treatment.  Therefore, the number of performed lab tests and relative findings  significantly varies among the considered set of patients (from only three to hundreds).

Moreover, the lab tests and X-ray exams are not performed at a regular frequency due, e.g., to the different kinds and timing of the relative procedures, the need of different resources (X-Ray machines, lab equipments, technical staff, etc.), or to the different severity of the health conditions of the patients. For example, in our data we see that a patient can be tested for PCR everyday and not be subject to a Ferritin exam for two weeks. This leads to the need of handling the issues \textit{missing values} and \textit{outdated values}. When we consider a snapshot of a patient at a certain day, we have a missing value for a lab test (or X-ray) feature if that test (X-ray) has not been performed. We have an outdated value for a feature if the corresponding lab test (X-ray) was performed several days earlier: since in the meanwhile the disease has progressed, the findings of the lab test could be inconsistent with the current conditions of the patient, and so they could mislead the prediction.
 

Data quality issues  arise especially patients hospitalised in the period of the highest emergency, when several hundreds of patients were in the hospital at the same time.

\subsection{Concept Drift}
\label{sec:drif}
\begin{figure*}
    \centering
    \includegraphics[scale=0.5]{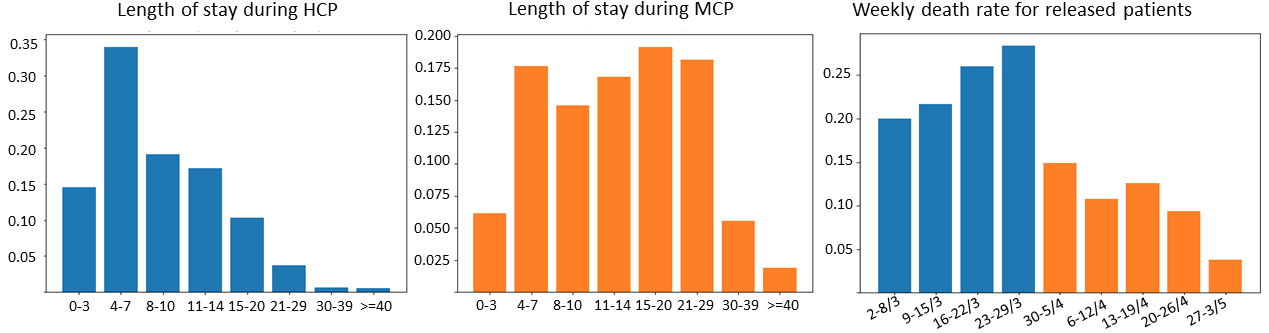}
    \vspace{-2mm}
    \caption{Length of stay in hospital (left) and weekly death rate histograms for the High Contagion Phase (in blue) and for the Moderate Contagion Phase (in orange). On the x-axis, for the length of stay we indicate the range of days, for the death rate we indicate the week when the patient was released. On the y-axis we indicate the percentage of patients.}
    \label{fig:los_ed_md}
\end{figure*}

An examination of the data available for our cohort of patients revealed that their prognostic risk is influenced by multiple factors, such as the number of the patients currently 
hospitalised
and the consequent availability of ICU beds or other resources, the experimentation of new therapies, and the increase of the clinical knowledge.

In machine learning, this change of data distribution is known as \textit{concept drift} \cite{concept_drift, dataset_shift}. A classical method to deal with this problem is training the algorithm using only a subset of samples, depending on the data distribution that we are considering  \cite{concept_drift, rakitianskaia2012training}.

For this reason, we divided the considered set of patients into two groups: the \textbf{High Contagion Phase (HCP)} group of patients, which is composed by the patients admitted during the last weeks of February and the first weeks of March (the most critical period of the pandemic outbreak in Italy) and the \textbf{Moderate Contagion Phase (MCP)} group of patients, which is composed by the patients admitted from the last decade of March to the end of April.

The main differences between these groups of patients are:
\begin{enumerate}
    \item the mortality rate of the  HCP patients is about twice the mortality rate of the MCP patients;
    
    \item in HCP patients the median value of the hospitalisation period is 8 days, while in MCP patients is 14 days. Further details are given in Figure \ref{fig:los_ed_md};
    \item for many of the considered lab test, the mortality rate associated with having values in a particular range  significantly changes in the two groups. For example, in HCP patients the mortality rate for the patients which had a PCR value 10 times above the normal range is 40.1\%, while in MCP patients it is 21.1\%.
\end{enumerate}

These differences clearly indicate that the 
data in the HCP and MCP groups represent different target (concept) functions;
therefore predicting  mortality during the high infection phase and during the moderate phase can be considered as two different tasks. If we had only the patients hospitalised during the high infection phase, using these data for training an algorithm that predicts the mortality during the moderate phase would lead to many errors. 

%
%
In our case, we generated two different systems, one for each of the two groups of patients. 
We are currently investigating ways to automatically select the set of patients for training starting from the latest ones, and keeping the less recent ones until we find significant changes in the mortality rate or in the data distribution.

\section{Datasets for Training and Testing}
The main task of our work is to provide survival/death predictions at different days of the patient hospitalisation, according to the current patient conditions reflected by the available lab findings and X-ray scores. In this section we describe the specific extracted features and the (training and testing) datasets that we built for this purpose. 

\subsection{Pre-processing and Feature Extraction}
The issues presented in Section \ref{sec:issues} compel us to a robust pre-processing phase with the goal of extracting features in order to summarize the patients conditions and process them by a machine learning algorithm. The pre-processing is applied to both HCP and MCP data.

Given that we have no information about the survival or the decease of a patient after a transfer (which can be due to limited availability of beds or ICU places), we exclude from our training and test set the 142 patients which were admitted in \textit{Spedali Civili di Brescia} and then transferred to another hospital. 
However, the 74 patients who were transferred to a rehabilitation center can be considered not at risk of death; therefore we include them in our datasets and consider the transferred patients as released alive.

\subsubsection{Patient Snapshot and Feature Engineering}
In order to provide a prediction for a patient at different hospitalisation times, we introduced the concept of \textbf{patient snapshot} to represent the patient health conditions at a given day.

In this snapshot,
for each lab test of Table \ref{tab:lab_values},  we consider its most recent value. In the ideal case, we should know the lab test findings at every day.
However, as explained in Section \ref{sec:issues}, in a real-world context the situation is very different. For example, in our data if we consider to take a snapshot of a patient 14 days after the admission into the hospital, we have cases with very recent values of PCR, LDH or WBC (obtained one or a few days before), very old values for Fibrinogen or Troponin-T (obtained the first day of the hospitalisation) and even no value for Ferritin.

Given the difficulty to set a predefined threshold that separates recent and old values of the lab tests (e.g., for Fibrinogen and Troponin-T), we choose to always use the most recent value, even if it could be outdated. In order to allow the learning algorithm to capture that a value may not be significant to represent the current status of the patient (because too old), we introduce a feature called \textbf{ageing} for each test finding. If a lab test has been performed at a day $d_0$, and the snapshot of a patient is taken at day $d_1$, the ageing is defined as the number of days between $d_1$ and $d_0$. If there is no available value for a lab test, its ageing is considered a missing value.

A patient snapshot can contain the values of the lab test findings in two forms: either \textbf{numerical}, in which we report the value itself, or \textbf{categorical}, in which the value is transformed into an integer number expressing the gravity of the test finding within a partition of the possible real values. This partition is based on the range of values for normal conditions and on how the test values are distributed over the data of all patients. For example, we divide the D-Dimer vales into 6 categories: the normal range, up to 2 times the maximum value of the normal range, up to 4 times, 6 times, 10 times and over 10 times. The categorical form could help the algorithm to have a clearer understanding of the data and improve performance.

Monitoring the conditions of a patient means knowing not only the patient status at a specific time, but also how the conditions evolve during the hospitalisation. For this purpose, we introduce a feature called \textbf{trend} that is defined as follows:
\begin{itemize}
\item[]
For each lab test, if there is no available value for a lab test or if the patient has not performed the lab test at least two times, the trend is a \textit{missing value}. Otherwise, 
given the values $v_1$ and $v_2$ of the findings for the lab test performed at days $d_1$ and $d_2$ and a threshold $T$ that we set to 15\% of $v_1$, if $v_2 > (1 + T) * v_1$, then the trend is \textit{increasing}, while  if $v_2 < (1 - T) * v_1$ the trend is \textit{decreasing}; otherwise the trend is \textit{stable}.
\end{itemize}

We distinguish two types of trends: the \textbf{start trend}, that uses the distance between the most recent value and the first available value, and the \textbf{last trend}, that uses the distance between the last one and the penultimate one. We are currently investigating techniques for including more than two values in the trend calculation.

To summarize, for each lab test in a patient snapshot, we have the most recent finding and the relative ageing and trend, as well as 
the static features {\bf age} and {\bf sex}.

\subsection{Training and Test Sets Generation}
In this section we describe how we generated the training and test sets for the purpose of 
predicting, at different days from the start of the patient hospitalization, the final outcome of her/his stay.

First, for both the HCP and MCP sets, we used stratified sampling for selecting \textit{80\% of the patients for training} the models and \textit{20\% for testing} them.
Then, we created specific training and test sets for each element in a sequence of times when the model is used to make the prediction\footnote{While we chose 2, 4, 6, 8, 10 days after the hospitalisation, plus the day before the patient release, of course other sequences could be considered.}:
\begin{itemize}
    \item \textbf{2 days} of hospitalisation. We include all the patients' snapshots containing the first values for each lab test conducted in the first two days after the hospital admission. Note that if a patient has performed a lab test more than once in the first two days, the snapshot will consider the oldest value. In fact, the purpose of the model we want to build is to provide the prediction as soon as possible, with the first information available. Furthermore, in these snapshots the ageing and trend values are not included.
    \item \textbf{4 days} and \textbf{6 days} of hospitalisation. In these cases, the corresponding snapshots also contain the ageing and trend features, and the lab values will be the most recent ones in the available data. Given that only a few days passed after admission, we consider the {\it start} trend.
    \item \textbf{8 days} and \textbf{10 days} of hospitalisation. The procedure of creating the corresponding snapshots is the same as for the snapshots of 4 days and 6 days cases, except that we consider the {\it last} trend instead of the start trend.
    \item \textbf{End day} (the last day before the patient is released or the patience decease).
    In this case, for each lab test the snapshot includes both the start trend and the last trend features. 

\end{itemize}


It is important to observe, that while the datasets of the latter days will contain more information about the single patients (more lab tests findings, less missing values), the overall number of patients in the datasets decreases with the prediction day increase. 
This is due to the fact that more patients are released or die within longer periods of hospitalisation, and therefore such patients are 
not included in the corresponding datasets.
%

Finally, note that the splitting between training and testing of the data is done only once considering all patients. 
Thus if, for instance, a patient belongs to the training set of 2 days, then it does not belong to the test set of the following days.


\section{Machine Learning Algorithms}
In this section we briefly describe the machine learning algorithms used in our prognosis prediction system.
\label{sec:algo}

\subsection{Classification Algorithms}

\subsubsection*{Decision trees}
Decision Trees \cite{Rokach:2008} are one of the most popular learning methods for solving classification tasks. In a decision tree, the root and each internal node provides a condition for splitting the training samples into two subsets depending on whether the condition holds for a sample or not.
In our context, for each numerical feature $f$, a candidate splitting condition is $f\leq C$, where $C$ is called \textit{cut point}. 
The final splitting condition is chosen by finding the $f$ and $C$  providing the best split according to one of some possible measures like Information Gain, Entropy index or Gini index.

A subset of samples at a tree node can either be split again by further feature conditions forming a new \textit{internal node}, or form a \textit{leaf node} labelled with a specific classification (prediction) value; in our application domain
the label is either the \textit{alive} class or the \textit{dead} class. 
Let us consider a decision tree
with a leaf node $l$ and a subset $S$ of associated {\it training} samples. A {\it test} instance $X$ that reaches $l$ from the root tree, is classified (predicted) $y$ with probability 
\[
 P(y|X) = \frac{TP}{TP + FP}
\]
where $TP$ (True Positives) is the number of training samples in $S$ that have class value $y$, and $FP$ (False Positives) is the number of samples in $S$ that don't have class value $y$ \cite{chawla2006evaluating}.
Given that in our task we have only two classes ($y$ and $\overline{y}$), 
$P(\overline{y}|X)= 1-P(y|X)$. 
The classification outcome of a decision tree for$X$ is the class value with the highest probability.

\subsubsection*{Random Forests}
Random Forests (RF) \cite{breiman2001random} is an ensemble learning method \cite{Zhou:2012} that builds a number of decision trees at training time. For building each individual tree of the random forest, a randomly chosen subset of the data features is used. While, in the standard implementation of random forests the final classification label is provided using the statistical mode of the class values predicted by each individual tree, in the well-known tool Scikit-Learn \cite{scikit-learn} that we used for our system implementation, the probability of the classification output is obtained by averaging the probabilities provided by all trees. Hence, given a random forest with $n$ decision trees, a class (prediction) value $y$ is assigned to an instance $X$  with the following probability:
\[
    P(y|X) = \frac{\sum_{i=1}^n P_i(y|X)}{n}.
\]

\subsubsection*{Extra Trees}
Extremely Randomized Trees (Extra Trees or ET) \cite{geurts2006extremely} are another ensemble learning method based on decision trees. The main differences between Extra Trees and Random Forests are:
\begin{itemize}
    \item 
    In the original description of Extra Trees \cite{geurts2006extremely} each tree is built using the entire training dataset. However in most implementations of Extra Trees, including Scikit-Learn \cite{scikit-learn}, the decision trees are built exactly as in Random Forests.
    \item In standard decision trees and Random Forests, the cut point is chosen by first computing the optimal cut point for each feature, and then choosing the best feature for branching the tree; while in Extra Trees, first we randomly choose $k$ features and then, for each chosen feature $f$, the algorithm randomly selects a cut point $C_f$ in the range of the possible $f$ values. This generates a set of $k$ couples $\{(f_i,C_i) \,\,| \,\,i = 1, \dots, k\}$. Then, the algorithm compares the splits generated by each couple (e.g., under split test $f_i \le c_i$) to select the best one using a split quality measure such as the Gini Index or others.

\end{itemize}
The probability $P(y|X)$ of assigning a class value $y$ to an instance $X$ is computed as in Random Forests (see equation above).

\subsection{Hyperparameter Search}
\label{sec:search}
Most machine learning algorithms have several hyperparameters to tune such as, for instance, in a Random Forest the number of decision trees to create and their maximum depth. Since in our application handling the missing values is an important issue, we also used a hyperparameter for this with three possible settings: a missing value is set to either the average value, the median value or a special constant (-1).


In order to find the best performing configuration of the hyperparameters, we used the Random Search optimization approach \cite{bergstra2012random}, which consists of the following main steps:
\begin{enumerate}
    \item We divide our training sets into $k$ folds, with either $k=10$ or $k=5$, depending on the dimension of the considered dataset.
    \item For each randomly selected combination of hyperparameters, we run the learning algorithm in $k$-fold cross validation.
    \item For each fold, we evaluate the performance of the algorithm with that configuration using the Macro F-$\beta$ score metric and $\beta = 2$. 
    The $F\textnormal{-}\beta$ score is the weighted harmonic mean of precision and recall measures. The $\beta$ parameter indicates how many times the recall is more important with respect to the precision:
    \vspace{-0.1cm}
    \[
        F\textnormal{-}\beta = (1+\beta^2)*\frac{precision + recall}{\beta^2*precision + recall}
    \]
       \vspace{-0.1cm}
    We choose $\beta = 2$ in order to give particular importance to false negatives, i.e. those patients which our system could not identify as at death risk.
    Given that we can compute the F2-score both for both the \textit{alive} class and the \textit{dead} class, we considered the Macro F2-Score, which is the arithmetic mean of the scores for the two classes.

\item The overall evaluation score of the $k$-fold cross validation  for a configuration of the parameters is obtained  by averaging the scores obtained for each fold.
\item  The hyperparameter configuration with the best overall score is selected.
\end{enumerate}

\subsection{Handling Prediction Uncertainty}
\begin{figure}[h]
\removelatexerror
\begin{algorithm}[H]
    \SetAlgorithmName{\sc FindUncertainThreshold}{\sc FindUncertainThreshold}{\sc FindUncertainThreshold}
    \caption{Algorithm for computing, during the training phase, an optimised prediction  threshold under which the model labels an instance as uncertain.}
    \KwIn{
        \begin{itemize}[label={--}]
            \item $L$ array of labels ($alive$ or $dead$) $l_{i}$ with $l[i]$ label of the 
             sample $i$ of the validation data (fold);
            \item $P = [p_i = (p_{alive}, p_{dead})_i \mid i \text{ is the sample index in val. set}]$;
            \item $max\_u$ the maximum percentage of the samples in the 
            validation set that can be labeled as uncertain (not predictable);
            \item $n$ the maximum number of thresholds to try;
            \item $EvaluateScore$ the score function to maximize by dropping the uncertain samples;
        \end{itemize}
    }
    \KwOut{A pair $(v, th)$ where $v$ is the score function value after dropping the uncertain samples and $th$ the optimized threshold value.}
    $L_{pred} \gets$ array of labels such that 
    $L_{pred}[i]$ is the predicted label (the label with highest probability) of the 
    val. sample $i$\;
    $P_{max} \gets [max(p_{alive}, p_{dead})_i \mid (p_{alive}, p_{dead})_i \in P] $\;
    $v \gets EvaluateScore(L, L_{pred})$\;
    $th \gets$ min value in $P_{max}$\;
    $\delta \gets$ [(max value in $P_{max}$) $-$ (min value in $P_{max}$)]$/n$\;
    \For{$i \gets 0$ to $n-1$}{
        $th' \gets $ min value in $P_{max} + i \cdot \delta$\;
        
        $S \gets 
        \{i \, | i$ is id sample such that $P_{max}[i] > th' \}$
        
        $u \gets 1 - (|S|/|P_{max}|)$\;
        
        \lIf{$u \geq max\_u$}{
            \Return $(v, th)$}
        
        $L' \gets$  array of labels such that 
        $L[i]$ is the label of the 
        val. sample $i$ and $i \in S$\;

        $L'_{pred} \gets$  array of labels such that 
        $L_{pred}[i]$ is the predicted label of the 
        val. sample $i$ and $i \in S$\;
        
        $v' \gets EvaluateScore(L', L'_{pred})$\;
        \If{$v' > v$}{
            $th \gets th'$\;
            $v \gets v'$\;
        }
    }
\end{algorithm}
   \vspace{-0.1cm}
\caption{Pseudocode of algorithm \sc FindUncertainThreshold.}
\label{alg:score_without_uncertain}
\end{figure}
The output for an instance $X$ of every generate classification model is an array of two probabilities, $P(alive|X)$ and $P(dead|X)$, defined as described in Section \ref{sec:algo}.1.
We can see them as ``degrees of certainty'' of the prediction: the higher the probability is, the more reliable the prediction is.
Given the very sensitive nature of our task, the system discards potential predictions supported by a low probability. This is achieved using a {\it prediction threshold} under which the system considers the prediction \textbf{uncertain} (and the patient risk unpredictable).
%
%
%
Note that if we used a threshold value that is too high, many patients could be classified uncertain, and our model would be much less useful for clinical practice. To avoid this, at training time we impose a maximum percentage of samples that can can be considered uncertain (unpredictable), and we implemented this with a parameter, called  $max\_u$, that is given in input; for our experimental analysis we used $max\_u = 25\%$.

We designed an algorithm called {\sc FindUncertainThreshold} that is used in the training phase to decide the threshold and optimize the prediction performance on the training samples that pass it, under the $max\_u$ constraint.
%
The pseudocode of the algorithm is in Figure\,\ref{alg:score_without_uncertain}.

Given the original labels $L$ of the validation samples and their prediction probabilities $P$ derived by
the learning algorithm, 
{\sc FindUncertainThreshold} first computes: the predicted labels $L_{pred}$ (i.e., the class values with highest probabilities) and the relative $P_{max}$ probabilities;
the original score $v$ obtained using the input score function evaluating {\it all} samples;
an initial value of the threshold ($th$) defined as to the minimum  probability in $P_{max}$.

The next loop finds an optimal value of threshold $th$ and computes the score function for the validation set reduced to the 
validation samples with predicted labels that have probabilities above $th$. The considered threshold values are obtained by using the $\delta$-increments defined at lines 5 and 7.
First we compute the new threshold $th'$ increasing the current threshold by $\delta$, and then we derive the set $S$ of sample ids with prediction probabilities higher than $th'$.
Next we compute the percentage $u$ of samples that are labeled as uncertain using threshold $th'$. If $u \ge max\_u$, we can terminate returning the current best new score $v$ and the corresponding threshold value $th$ (a greater threshold value cannot lead to label as uncertain less samples than the returned $th$ value). 
Otherwise ($u < max\_u$), we compute the correct sample labels $L'$ and the predicted sample labels $L'_{pred}$ for the samples
identified by $S$,
and we compute the new score value $v'$ using $L'$ and $L'_{preds}$.
If $v'$ is a better score than $v$, we update both the threshold and the score values.

{\sc FindUncertainThreshold} is executed during the training phase.
In particular during the hyperparameter search, for each attempted hyperparamenter configuration, we compute through {\sc FindUncertainThreshold} an optimized threshold and the relative score function value. 
These two values are obtained by averaging the optimal thresholds and corresponding scores
over all folds of the cross validation for the attempted configuration.
The hyperparameter search returns the best configuration together with the relative (averaged) threshold.

\vspace{-3mm}
\section{Experimental Evaluation and Discussion}
\begin{figure}[t]
    \centering
    \includegraphics[scale=0.35]{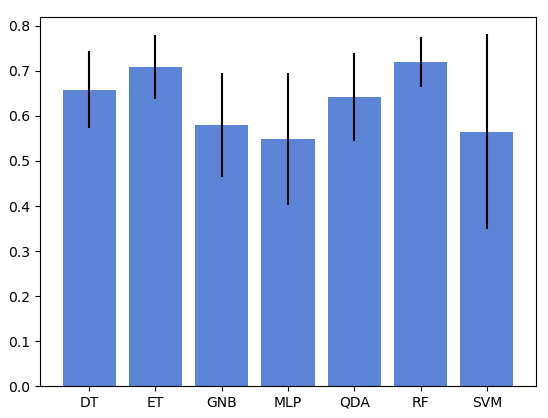}
    \vspace{-2mm}
    \caption{Average performance (F2 score) of seven machine learning algorithms for the HCP datasets. The line over the bar represents the standard deviation.}
    \label{fig:comparison}
\end{figure}
\begin{figure*}[h]
    \centering
    \includegraphics[scale=0.35]{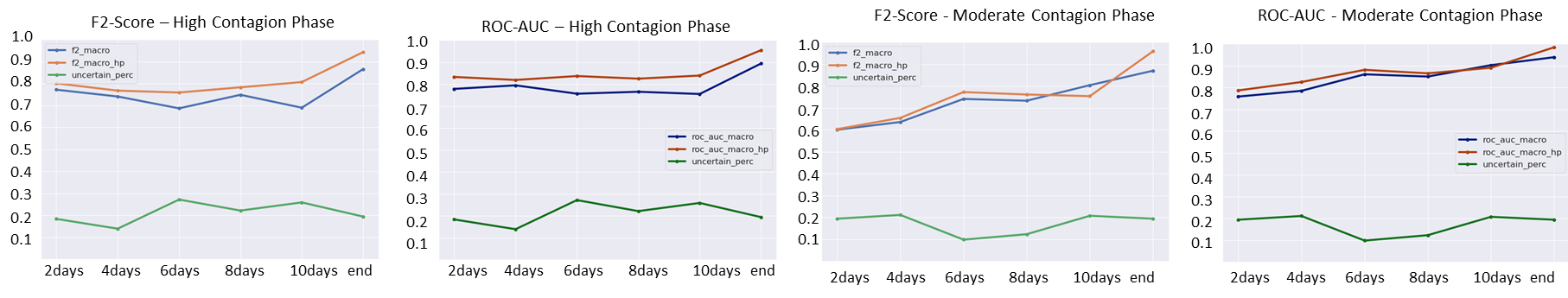}
   \vspace{-2mm}
    \caption{Graphical representation of the prediction performance  
    (F2 and ROC-AUC scores) over hospitalisation time  for HCP and MCP.}
    \label{fig:results}
\end{figure*}
\begin{table*}[h]
    \centering
    \begin{tabular}{l| c | c| c | c | c | c ||| l | c| c | c | c | c | c }
         HCP data   & F2 & ROC & F2-U & ROC-U & \% Unc & Model & MCP data & F2 & ROC & F2-U & ROC-U & \% Unc & Model\\ \hline
         2days & 77.1 & 77.8 & 80.1 & 83.3 & 18.3 & ET-C & 2days & 60.0 & 75.4 & 61.0 & 78.1 & 13.9 & ET-N \\
         4days & 74.1 & 79.4 & 76.7 & 81.9 & 13.8 & RF-N & 4days & 63.5 & 78.5 & 65.4 & 82.4 & 21.1 & RF-N \\
         6days & 68.7 & 75.6 & 75.9 & 83.6 & 26.1 & RF-N & 6days & 74.1 & 86.0 & 77.2 & 88.1 & 9.8 & ET-N \\
         8days & 74.8 & 76.5 & 78.2 & 82.5 & 22.1 & ET-C & 8days & 73.2 & 85.0 & 76.1 & 86.5 & 12.3 & ET-N \\
         10days & 68.9 & 75.5 & 80.6 & 83.9 & 24.8 & RF-C & 10days & 80.4 & 90.2 & 75.3 & 89.0 & 12.7 & ET-N \\ 
         end & 86.6 & 89.4 & 94.3 & 95.5 & 19.3 & RF-C & end & 86.9 & 93.9 & 95.8 & 98.4 & 19.4 & RF-N \\
    \end{tabular}
   \vspace{-1mm}
    \caption{Predictive performance for the High Contagion Phase (HCP, left) and the Moderate Contagion Phase (MCP, right) in terms of F2 and ROC-AUC scores considering all instances in the test set 
    (columns F2 and ROC) and omitting the instances classified uncertain (columns F2-U and ROC-U).
      The percentages of instances that the system classifies uncertain are in the column \% Unc. Column Model indicates the method selected for generating the model; ET stands for Extra Trees, RF for Random Forest, C for categorical and N for numerical.}
    \label{tab:best_results}
\end{table*}

In this section, we evaluate the performance the of the machine learning models that we built. Our system was implemented using the Scikit-Learn \cite{scikit-learn} library for Python, and the experimental tests were conducted using a Intel(R) Xeon(R) Gold 6140M CPU @ 2.30GHz.

The performance of the learning algorithms with the relative optimized hyperparameters was evaluated using the test set in terms of F2 score and ROC-AUC score. The second metric is defined as the area under the Receiver Operating Characteristic curve, which plots the true positive rate against the false positive rate, and it takes also into account the probability that the predictive system produces false positives (i.e. false alarms). This metric is a standard method for evaluating medical tests and risk models \cite{grunkemeier2001receiver, hajian2013receiver}.

In a preliminary study we examined various machine learning approaches and we compared their average performances over the HCP datasets. Figure  \ref{fig:comparison} shows a summary of the relative performance in terms of F2 score. We considered
 Decision Trees \cite{Rokach:2008}, ExtraTrees (ET) \cite{geurts2006extremely}, Gaussian Naive Bayes \cite{naivebayes}, Multilayer Perceptron with two layers (MLP) \cite{haykin1994neural}, Quadratic Discriminant Analysis \cite{qda}, Random Forests (RF) \cite{breiman2001random} and Support Vector Machines \cite{svm}.
 The best performance was obtained with RF and ET. NN and SVM performed much worse and with a much higher variability over the datasets,
 probably related to the missing values and the scarcity of data.
 For the MCP datasets the relative performance was similar.
 Given the observed better performance of RF and ET, we focused the evaluation of our system on these learning algorithms 
 
Regarding the training time, including the hyperparamenter search over 4096 random configurations and the optimization of the uncertainty threshold, for any specific dataset (e.g., the MCP numerical dataset for 2 days),
the overall training time is between 20 and 30 minutes.
Therefore, we can build all the four most promising models generated by RF and ET
using the numerical version (RF-N, TC-N) or the categorical version (RF-C, ET-C) of the data set in less than two hours, and then select the best performing model among them.

 It is also worth to note that in our system the models for predicting the prognostic risk at different days are completely independent from each other, and so we can consider prediction tasks at different days as different tasks. 

In Figure \ref{fig:results} and in Table \ref{tab:best_results} we show the  performances of our system at each considered day for both the High Contagion Phase and the Moderate Contagion Phase. As we can see, we obtain promising results in terms of F2 score for an early evaluation of the risk during the HCP (with score 77.1\% at day 2), while we encounter some problems at the 6th and 10th days. For the MCP datasets, the system performs better at the latter days, in particular for the 10th day F2 is 80.4\% and ROC-AUC is 90.2\%.
For HCP, both RF and ET obtain good results in both the numerical and categorical versions of the datasets. Instead, for MCP using the categorical datasets does not give good performance, and we do not observe an improvement for the latter prediction days (the F2 score is always below the 70\%).

In all but one case, the models using the uncertain threshold increase the performance in terms of both F2 and ROC-AUC scores. In particular, in the most problematic cases of HCP, such as for the 6-days and 10-days datasets, the prediction performance improves in terms of F2 by over than 7 points. The improvement is less significant for MCP.

Note that, while the threshold value under which the system labels an instance (patient risk) as uncertain is derived at training time imposing a maximum percentage of uncertain samples (we used 25\%), there is no formal guarantee that this percentage limit is satisfied for test set. However, in most cases the percentage of uncertain test samples (indicated with \% Unc in Table \ref{tab:best_results}) is much below the limit imposed during training, expect for the test set of the 6th day in HCP, where the unpredicted  (labelled as uncertain) patients are
26.1\%.
The performance for the ``end'' dataset is good for both HCP and MCP even without omitting the uncertain patients (F2 score 86.6\% for HCP, and F2 score 86.9\% for MCP).

Figure \ref{fig:results} gives graphical pictures comparing the performance of our system for HCP and MCP
in terms of F2 and ROC-AUC. The performance behaviour over time significantly differs in the two contagion periods, reflecting the concept drift we discussed in Section \ref{sec:drif}. 
%
For HCP, considering the results without omitting the uncertain test instances (blue curves), the performance prediction is very good at the 2nd day and it decreases at the 6th and 10th days. Instead, for MCP the performance improves over time, reaching 90.2\%
in terms of ROC-AUC at the 10th day, as also reported in Table \ref{tab:best_results}.
This is due to several factors:
\vspace{-0.2cm}
\begin{itemize}
    \item MCP includes patients that have hospitalisation periods much longer than the patients in HCP, which can make more difficult to predict the mortality risk for
    some patients with only a few days of hospitalisation;
    \item on the contrary, in HCP half of the patients stayed in hospital for less than 8 days. This decreases significantly the size of the 8-days and 10-days training sets, which contain respectively only 431 and 339 patients. The lack of training data in these datasets is only partially compensated by the increase of the lab tests for a single patient in the datasets;
    \item as described in Section \ref{sec:drif}, the MCP patients are much more unbalanced (with only 11\% deceased patients) than the HCP patients, and this increases the difficulty of learning an high performing model \cite{krawczyk2016learning}.
\end{itemize}

\begin{figure}[h]
    \vspace{-2mm}
    \centering
    \includegraphics[scale=0.33]{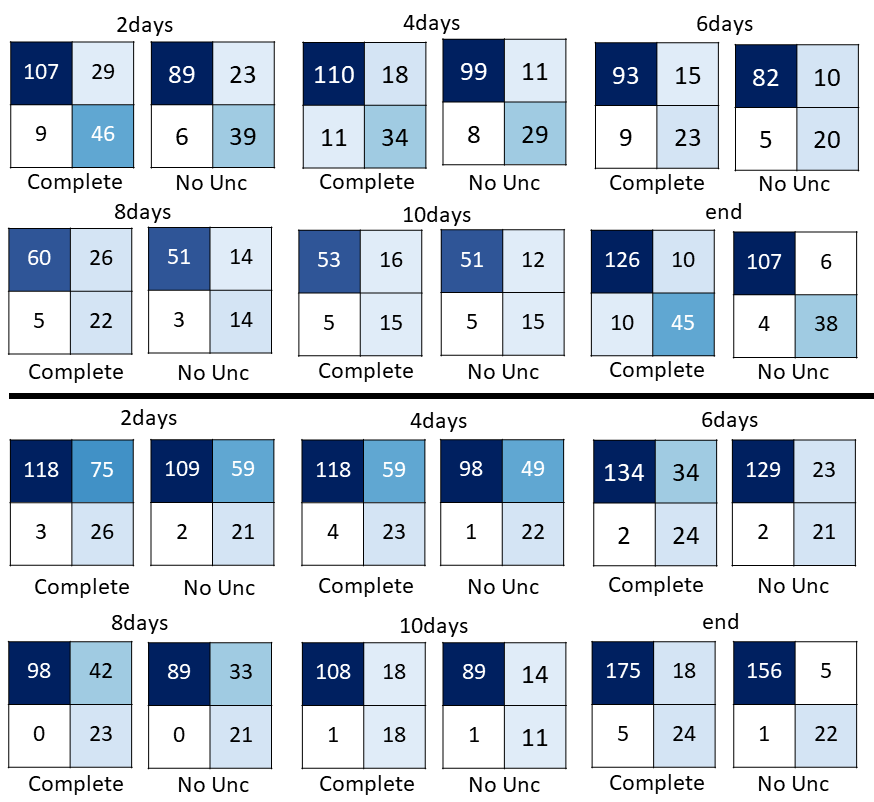}
    \vspace{-1mm}
    \caption{Confusion matrices for datasets
    HCP (above the line) and MCP (below the line) at different days
    with {\it dead-alive} predictions for all patients (Complete) and omitting patients classified uncertain  (No Unc). For each matrix of 4 numbers,  on the main diagonal we have the correct predictions
    (\textit{alive} class on the top-left corner and \textit{dead} class on the bottom-right corner); on the anti-diagonal, we have the incorrect predictions 
    (false positives and on the top-right corner and the false negatives on the bottom-left corner).
    }
    \label{fig:matrices_vert}
\end{figure}

Figure \ref{fig:matrices_vert} shows the confusion matrices for the test sets generated using our  predictive models. Above the line we have the HCP datasets and below the MCP datasets. Despite the training phase was optimised (through the use of the F2 metric) to avoid false negatives, for the HCP datasets there are several false negatives (bottom-left of the matrices). This can be explained by the scarcity of lab test and X-ray data in the HCP data that affects prediction.

However, false negatives are significantly reduced with the models that can classify a
patient as uncertain.
For example, at day 6, the system classifies as uncertain 4 patients who otherwise would be false negatives. Moreover, when there are less false negatives, such as at days 8 and 10,  classifying  some patients as uncertain helps to also avoid false positives and so to generate less false alarms. 

Remarkably, especially for the MCP datsets, we have very few false negatives even at the early days, which is quite important in our application context.
On the other hand, especially for days 2 and 4, our system produces many false positives.
This type of error is reduced in the models with uncertain patients up to 
only 5 false alarms for the end dataset (e.g., at day 2 
we avoid 16 false positives.)
\vspace{-0.2cm}
\section{Conclusions and Future Work}
\vspace{-2mm}
We have presented a system for predicting the prognosis of Covid-19 patients focusing on the death risk. We built and engineered some datasets from lab test and X-ray data of more than 2000 patients in an hospital in northern Italy that was severely hit by Covid-19. Our predictive system uses a collection of machine learning algorithms and a new method for setting, at training time, an uncertain threshold for prediction that helps to significantly reduce the prediction errors.

Overall, the experimental results are quite promising, and show that our system often obtains high ROC-AUC scores.
The observed predictive performance is especially good in terms of false negatives (patients erroneously predicted  survivor), that are very few. This gives a predictive test for patient survival with very good 
specificity
in particular when the system can classify a patient as uncertain.




On the other hand, in terms of false positives, there is room for significant improvements. We are confident that the availability of more information, such as patient comorbidities or clinical treatments, will help to improve performance, reducing the number of both false positives and (few) false negatives.

For future work we plan to extend our datasets with more information (both additional features and patients), to consider further methods for dealing with the observed concept drift
and to address other prediction tasks such as the duration of the hospitalisation or the need of ICU beds and critical hospital resources. Moreover, we are analyzing the importance of the features used in our models, and we intend to investigate additional learning techniques.


  \vspace{0.4cm}
\noindent
{\large \bf Acnowledgements}.
The work of the first author has been supported by Fondazione Garda Valley.
\bibliography{ecai}
\end{document}